\documentclass[10pt,twocolumn,letterpaper]{article}

\usepackage{iccv}
\usepackage{times}
\usepackage{epsfig}
\usepackage{graphicx}
\usepackage{amsmath}
\usepackage{amssymb}
\usepackage{cite}
\usepackage{gensymb}
\usepackage{textcomp}
\usepackage{ragged2e} 
\usepackage{booktabs,makecell, multirow, tabularx}
\usepackage[table]{xcolor}
\usepackage{caption}
\usepackage{placeins}%\FloatBarrier
\usepackage[breaklinks=true,bookmarks=false]{hyperref}
\usepackage{flushend}
\hypersetup{
colorlinks=true
}

\usepackage[breaklinks=true,bookmarks=false]{hyperref}
\usepackage{cleveref}

\iccvfinalcopy 

\def\iccvPaperID{****}

\ificcvfinal\fi

\def\authorBlock{
    Shibei Meng\textsuperscript{1}\thanks{Equal contribution}~,
    Yang Fu\textsuperscript{1}\footnotemark[1]~,
    Saihui Hou\textsuperscript{1,2}\thanks{Corresponding Author}~,
    Chunshui Cao\textsuperscript{2},
    Xu Liu\textsuperscript{2},
    Yongzhen Huang\textsuperscript{1,2}\footnotemark[2]  \\
    \textsuperscript{1~}School of Artificial Intelligence, Beijing Normal University~
    \textsuperscript{2~}WATRIX.AI \\
    {
    \tt\small \{mengshibei, yangfu\}@mail.bnu.edu.cn,
    
    \{chunshui.cao, xu.liu\}@watrix.ai
      
     }
    \\
     {
    \tt\small 
      \{housaihui, huangyongzhen\}@bnu.edu.cn
     }
}

\begin{document}

\crefname{equation}{Eq.}{Eqs.}
\crefname{figure}{Figure}{Figures}
\crefname{table}{Table}{Tables}
%%%%%%%%% TITLE
\title{FastPoseGait: A Toolbox and Benchmark for Efficient Pose-based Gait Recognition}
\author{\authorBlock}

\maketitle

\ificcvfinal\thispagestyle{empty}\fi

%%%%%%%%% ABSTRACT
\begin{abstract}

    We present FastPoseGait, an open-source toolbox for pose-based gait recognition based on PyTorch. Our toolbox supports a set of cutting-edge pose-based gait recognition algorithms and a variety of related benchmarks. Unlike other pose-based projects that focus on a single algorithm, FastPoseGait integrates several state-of-the-art (SOTA) algorithms under a unified framework, incorporating both the latest advancements and best practices to ease the comparison of effectiveness and efficiency. In addition, to promote future research on pose-based gait recognition, we provide numerous pre-trained models and detailed benchmark results, which offer valuable insights and serve as a reference for further investigations.
    By leveraging the highly modular structure and diverse methods offered by FastPoseGait, researchers can quickly delve into pose-based gait recognition and promote development in the field.
    In this paper, we outline various features of this toolbox, aiming that our toolbox and benchmarks can further foster collaboration, facilitate reproducibility, and encourage the development of innovative algorithms for pose-based gait recognition.
    FastPoseGait is available at \url{https://github.com//BNU-IVC/FastPoseGait} and is actively maintained. We will continue updating this report as we add new features.

\end{abstract}

%%%%%%%%% BODY TEXT
%-------------------------------------------------------------------------
\vspace{-0.4cm}
\section{Introduction}

Pose-based gait recognition~\cite{liao2017pose,liao2020model,teepe2021gaitgraph,an2020performance,teepe2022towards,zhang2022spatial} aims to identify individuals based on their gait patterns captured through human pose sequences, even from a distance. Unlike the silhouette modality~\cite{chao2019gaitset,fan2020gaitpart,hou2020gait,hou2022gait,lin2021gait,liang2022gaitedge}, poses offer keypoint representations that are more robust to changes in appearance. As a result, pose-based gait recognition has gained significant interest in recent years. Several algorithms have been developed to achieve pose-based gait recognition, including methods based on Graph Convolutional Networks (GCN)~\cite{teepe2021gaitgraph,teepe2022towards} and Transformer~\cite{zhang2022spatial}.

GaitGraph~\cite{teepe2021gaitgraph} was the pioneering work that introduced Graph Convolutional Networks (GCN) ~\cite{song2020stronger} into pose-based gait recognition. This novel approach demonstrated a notable improvement in performance compared to previous methods~\cite{liao2020model,an2020performance}. Building upon the success of GaitGraph, GaitGraph2~\cite{teepe2022towards} was presented to further enhance the performance of pose-based methods, particularly on the large dataset of OUMVLP-Pose~\cite{an2020performance}. GaitTR~\cite{zhang2022spatial} introduced the use of Transformer~\cite{plizzari2021spatial} to improve the performance even further by leveraging the global relationship modeling capabilities of self-attention. For instance, on the CASIA-B benchmark,  the rank-1 accuracy has already achieved up to 90.0\% in CL conditions.

However, these existing advanced methods are conducted on only one or two benchmarks, lacking a comprehensive comparison across indoor and outdoor scenarios and datasets of varying scales. Furthermore, none of the existing approaches have compared these architectures under a unified setting, hindering the ability to make direct performance assessments.
Toward the goal of addressing these limitations and offering a well-structured, highly modular and easily understandable codebase for researchers, we have built FastPoseGait. This toolbox offers several major features to enhance its usability and effectiveness:

\textbf{(1) Alignment of experimental settings.} FastPoseGait ensures a unified experimental setting, enabling quick reproduction of results and fair performance comparisons among different algorithms. This alignment facilitates efficient research and evaluation.

\textbf{(2) Modular architecture.} We decompose the gait recognition framework into distinct components, allowing users to easily construct a customized gait recognition algorithm by combining different modules.

\textbf{(3) Support of multiple algorithms and benchmarks. } FastPoseGait supports a range of cutting-edge gait recognition approaches and benchmarks with diverse research interests and different scales.

\textbf{(4) Availability of a large number of trained models and detailed results.} The toolbox contains a large-scale model zoo and corresponding comprehensive results on different datasets to help researchers acquire a comprehensive understanding of pose-based gait recognition in an efficient manner.

\textbf{(5) High efficiency.} The code described in this report is compatible with multi-GPU setups utilizing Distributed Data Parallel (DDP)\footnote{\url{https://pytorch.org/tutorials/intermediate/ddp_tutorial.html}}. Additionally, it supports model training with Auto Mixed Precision (AMP)\footnote{\url{https://pytorch.org/tutorials/recipes/recipes/amp_recipe.html}}. These features are implemented to accelerate the training process.

%-------------------------------------------------------------------------
\section{Supported Frameworks and Datasets}
\subsection{Supported Frameworks}
% par. Gaitgraph
\textbf{GaitGraph}~\cite{teepe2021gaitgraph} adopts the ResGCN model~\cite{song2020stronger} as a backbone, which is originally used in action recognition. By regarding the human pose as a graph and employing graph convolution operations, the proposed approach achieves notable enhancements in accuracy compared to earlier methods for pose-based gait recognition\cite{liao2017pose,liao2020model}. Furthermore, GaitGraph utilizes Supervised Contrastive Loss~\cite{khosla2020supervised} for optimization.

% par. GaitGraph2
\textbf{GaitGraph2}~\cite{teepe2022towards} is built upon GaitGraph and introduces a significant improvement by incorporating a multi-input mechanism. Additionally, certain modifications have been made to the number of layers in order to better accommodate the scale of datasets. Supervised Contrastive Loss~\cite{khosla2020supervised} is employed in this method with a much larger batch size compared with GaitGraph.

% par. GaitTR
\textbf{GaitTR}~\cite{zhang2022spatial} is the first algorithm to introduce Spatial Transformer~\cite{plizzari2021spatial} into pose-based gait recognition, enabling the establishment of comprehensive spatial relationships. This algorithm has achieved state-of-the-art performance on the CASIA-B benchmark~\cite{yu2006framework}. Triplet Loss~\cite{hermans2017defense} is adopted to optimize the training procedures.

\subsection{Supported Datasets}
% par. CASIA-B [SimCC HRNet]
\textbf{CASIA-B}~\cite{yu2006framework} consists of 124 subjects under three distinct walking conditions. 11 sequences of each subject on every walking condition are captured from different viewpoints.
HRNet~\cite{8953615} and SimCC~\cite{li2022simcc} are employed to extract pose data of 17 keypoints as shown in \cref{fig:graph} (a).

% par. OUMVLP-Pose [AlphaPose OpenPose]
\textbf{OUMVLP-Pose}~\cite{an2020performance} is built on the large-scale OUMVLP dataset~\cite{takemura2018multi}, which comprises 10,307 subjects from 14 viewpoints. OUMVLP-Pose provides 18-keypoint pose data shown in \cref{fig:graph} (b) extracted by AlphaPose~\cite{alphapose} and OpenPose~\cite{8765346}.

% par. GREW
\textbf{GREW}~\cite{zhu2021gait} is a large-scale dataset including 26,345 individuals captured by 882 cameras in real-world settings. GREW provides pose information comprising 17 keypoints, which is obtained using HRNet pose estimator~\cite{8953615}.

% par. Gait3D
\textbf{Gait3D}~\cite{zheng2022gait} is an in-the-wild dataset collected by 39 cameras from a fixed real-world environment, consisting of 4,000 subjects. Gait3D also includes pose information of 17 keypoints extracted by HRNet pose estimator~\cite{8953615}.

%-------------------------------------------------------------------------
\begin{figure*}
    \centering
    \includegraphics[width=\textwidth]{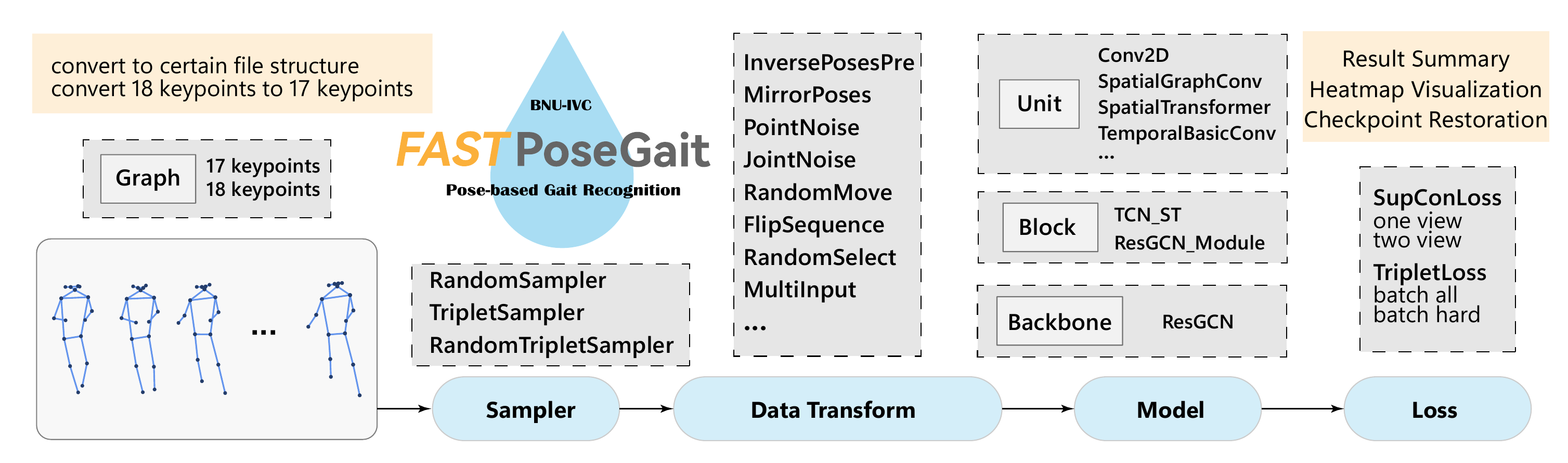}
    \caption{The Pipeline and Structure of FastPoseGait.}
    \label{fig:pipeline}
    \vspace{-0.3cm}
\end{figure*}

\section{Architecture}
\cref{fig:pipeline} illustrates the complete architecture of FastPoseGait, containing four distinct modules. Initially, \textbf{Sampler} selects sequences of skeletons, which are then pre-processed with the \textbf{Data Transform} module. \textbf{Model} encodes the inputs and is composed of various backbones. These backbones are comprised of blocks, which in turn consist of units with either residual connections or combinations of spatial and temporal convolutional layers. Different \textbf{Loss} functions are implemented to supervise the training process.
Furthermore, FastPoseGait incorporates specific scripts that guarantee the proper formatting of data, including the required file structures and keypoint orders.
TensorBoard\footnote{\url{https://www.tensorflow.org/tensorboard}} is utilized to conveniently visualize training logs and keypoint heatmaps, facilitating better record-keeping of the model optimization.
\subsection{Sampler}
\label{section:3.1}
\paragraph{Random Sampler}
Random sampler is utilized in GaitGraph~\cite{teepe2021gaitgraph} and GaitGraph2~\cite{teepe2022towards} which adopt the supervised contrastive loss~\cite{khosla2020supervised}. It randomly selects sequences from the whole dataset without considering the proportion of the positive and negative samples, allowing for a more diverse representation of data in one batch during training.
\vspace{-1.2em}
\paragraph{Triplet Sampler}
Triplet sampler is used in GaitTR~\cite{zhang2022spatial} which employs the triplet loss~\cite{hermans2017defense} to supervise the training process. The batch size of this sampler is defined as \((P,K)\), in which \(P\) refers to the number of subjects in a batch and \(K\) is the sequence number of every subject. This sampling strategy ensures a relatively balanced proportion of positive and negative samples.
\vspace{-1.2em}
\paragraph{Random Triplet Sampler}
Random Triplet Sampler is a \emph{trade-off} of Random Sampler and Triplet Sampler. To ensure the presence of both positive and negative pairs in a single batch, we propose the following constraints on \(P\) and \(K\):
\begin{equation}
    \label{eq:0}
    \left\{
    \begin{aligned}
         & {{P}}\geqslant 2, \\
         & {{K}}\geqslant 2, \\
         & PK+c=BatchSize,   \\
    \end{aligned}
    \right.
\end{equation}
where \(BatchSize\) is the total number of sequences in a batch. \(c\) denotes the remainder when \(BatchSize\) is divided by \(P\). After sampling \(P\) subjects with \(K\) sequences, \(c\) sequences are randomly selected from the remaining training set.
Additionally, \(P\) and \(K\) can be further limited within an upper bound based on the characteristics of the dataset. \eg, the training set of CASIA-B consists of 74 subjects and each subject is with 110 sequences, so \(P \leqslant 74\) and \(K \leqslant 110\).

\begin{figure}[!h]
    \centering
    \includegraphics[width=0.43\textwidth,height=0.37\textwidth]{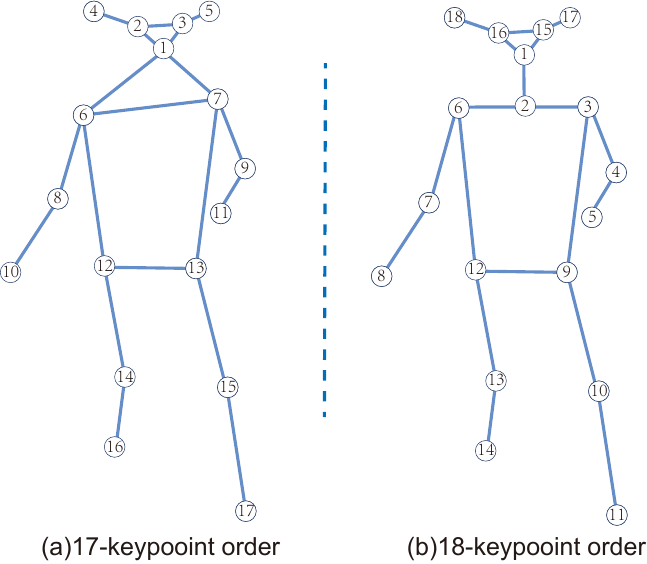}
    \caption{The illustration of keypoint orders and pre-defined connections. }
    \label{fig:graph}
    \vspace{-0.3cm}
\end{figure}

\subsection{Data Transform}
\paragraph{Spatial Augmentations}
We implement five commonly employed spatial augmentations in FastPoseGait. They can be divided into two categories:

(1) Spatial Flip.
(i) Pre-defined Inverse: It adopts the pre-defined symmetrical structure of keypoints to perform the left-right flipping operations. \eg, in \cref{fig:graph} (a), we exchange the coordinate values of $\textcircled{\scriptsize{6}}$ and $\textcircled{\scriptsize{7}}$, $\textcircled{\scriptsize{8}}$ and $\textcircled{\scriptsize{9}}$, \etc.
(ii) Mirror Poses: The \emph{centroid} coordinate values are acquired by calculating the mean of all keypoints in a skeleton graph. Then the symmetry points relative to these centroid values are obtained to mirror the skeletons.
%flip the skeletons.

(2) Spatial Noise.
This type of transformation introduces random Gaussian noise at different levels, namely the point level, temporal level, and the entire sequence level.
\vspace{-1.2em}
\paragraph{Sequence Augmentations}
We implement two kinds of sequence-level augmentations:
(1) Sequence Flip. It is to reverse the order of frames in a sequence.
(2) Random Selection: It is to randomly select a fixed number of ordered frames from the entire sequence.
\vspace{-1.2em}
\paragraph{Multi-Input} There are mainly four types of explicitly calculated input in action recognition~\cite{song2022constructing}, including joint, bone, angle and velocity. Various combinations are taken as input in different methods~\cite{teepe2022towards,zhang2022spatial}.

\subsection{Design of Model Framework}

\paragraph{Graph}
Graph is the priori knowledge of human skeletons. It defines adjacency matrices of skeleton topologies and symmetrical structures of human keypoints. FastPoseGait supports the 17-keypoint and 18-keypoint graphs based on different pose estimators as shown in \cref{fig:graph}.
The pre-defined graph plays a significant role in defining the relationships and connectivities between keypoints, facilitating effective graph convolutions.
\vspace{-1.2em}
\paragraph{Unit}
Unit is a collection of basic layers. Different methods incorporate distinct types of basic layers, which encompass both spatial and temporal layers. \eg, Graph Convolutional layer~\cite{yan2018spatial}, Spatial Transformer layer~\cite{plizzari2021spatial}, Temporal Convolutional layer~\cite{yan2018spatial}, \etc.
\vspace{-1.2em}
\paragraph{Block}
Block typically is comprised of one or two basic layers, representing a self-contained module repeated within a model backbone. It can consist of either a combination of a spatial and temporal layer or simply include a basic layer with residual connections.
The block serves as a fundamental building component in the model architecture, enabling the flexible  composition of these modules to capture complex spatial and temporal relationships.
\vspace{-1.2em}
\paragraph{Backbone}
Backbone is a model structure that is responsible for extracting discriminative features from transformed inputs. For instance, ResGCN~\cite{song2020stronger} composed of multiple blocks with residual connections is employed as the backbone in both GaitGraph and GaitGraph2.

\begin{figure}[]
    \centering
    \includegraphics[width=0.4\textwidth,height=0.19\textwidth]{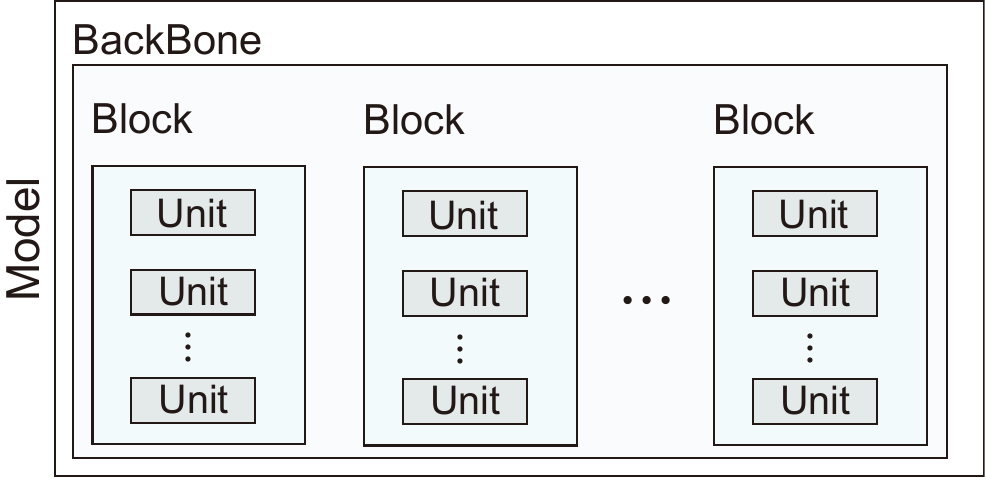}
    \caption{Framework of the model, illustrated with abstractions in FastPoseGait. }
    \label{fig:model}
    \vspace{-0.3cm}
\end{figure}

\vspace{-1.2em}
\paragraph{Model}
Model shown in \cref{fig:model}, serves as a method-specific definition of graphs, units, blocks and backbones. It outlines how these components are structured and combined to create an integrated framework to ensure consistency and effectiveness to achieve the desired objective.

%---------------------------

\subsection{Loss Function}
\paragraph{Triplet Loss}
Triplet Loss~\cite{hermans2017defense} is employed in GaitTR~\cite{zhang2022spatial} to decrease the intra-class distance and increase the inter-class distance of different identities. It can be divided into two types, one is named Batch All, in which all the triplets in a batch are considered:
\begin{equation}
    \label{eq:1}
    {{L}_{BA}}=\frac{1}{{{N}_{tri+}}}\overbrace{\underset{i=1}{\mathop{\overset{P}{\mathop{\sum }}\,}}\,\underset{a=1}{\mathop{\overset{K}{\mathop{\sum }}\,}}\,}^{\text{all anchors}}\overbrace{\underset{\begin{smallmatrix}
                p=1 \\
                p\ne a
            \end{smallmatrix}}{\mathop{\overset{K}{\mathop{\sum }}\,}}\,}^{\text{all pos}\text{.}}\overbrace{\underset{\begin{smallmatrix}
                j=1 \\
                j\ne i
            \end{smallmatrix}}{\mathop{\overset{P}{\mathop{\sum }}\,}}\,\underset{n=1}{\mathop{\overset{K}{\mathop{\sum }}\,}}\,}^{\text{all neg.}}{{[m+d_{i,a,p}^{j,a,n}]}},
\end{equation}
\begin{equation}
    \label{eq:2}
    d_{i,a,p}^{j,a,n}=\text{D}(f_{i}^{a},f_{i}^{p})-\text{D}(f_{i}^{a},f_{j}^{n}),
\end{equation}
where the total number of triplets in a batch is \(PK(PK-K)(K-1)\), \({N}_{tri+}\) refers to the number of triplets with non-zero loss values, \(m\) is the margin threshold between positive and negative pairs. \(\text{D}(\cdot)\) is the calculation of distance, \eg, Euclidean Distance, and \(f\) is final outputs of models.

The other type of triplet loss is named Batch Hard. For each anchor sequence within a batch, only the positive pair with the maximum distance is considered, as well as the negative pair with the minimum distance:
\begin{equation}
    \label{eq:3}
    {{L}_{BH}}=\frac{1}{{{N}_{tri+}}}\overbrace{\underset{i=1}{\mathop{\overset{P}{\mathop{\sum }}\,}}\,\underset{a=1}{\mathop{\overset{K}{\mathop{\sum }}\,}}\,}^{\text{all anchors}}[m+d_{i,a}],
\end{equation}
\begin{equation}
    \label{eq:4}
    d_{i,a}=\ \underbrace{ \mathop{max}\limits_{p=1...K}\ (\text{D}(f_{i}^{a},f_{i}^{p}))}_{\text{hardest positive}}
    -
    \overbrace{\mathop{min}\limits_{\begin{smallmatrix}
                j=1...P \\
                n=1...K \\
                j\ne i
            \end{smallmatrix}}\ (\text{D}(f_{i}^{a},f_{j}^{n}))}^{\text{
            hardest negative}},
\end{equation}
where the number of generated hard triplets of \(L_{BH}\) in a batch is \(PK\).

% par. Supervised Contrastive Loss
%TwoView / OneView
\vspace{-1.2em}
\paragraph{Supervised Contrastive Loss}
Supervised Contrastive Loss is originally proposed in~\cite{khosla2020supervised} to handle the case that more than one sample is known as positive due to the presence of labels. It offers improved performance by avoiding the requirement for explicit hard mining, which is a sensitive yet crucial aspect of various loss functions like triplet loss:
\begin{equation}
    \label{eq:3}
    {{L}_{sup }}=\sum\limits_{i\in I}{-\log \left\{ \frac{1}{\left| S(i) \right|}\sum\limits_{s\in S(i)}{\frac{\exp ({{f}_{i}}\cdot {{f}_{s}}/\tau )}{\sum\limits_{a\in A(i)}{\exp }({{f}_{i}}\cdot {{f}_{a}}/\tau )}} \right\}}.
\end{equation}
In one batch, \(I\) represents all the samples, while \(A(\cdot)\) represents all the samples except the one at index \(i\). \(S(\cdot)\) represents the positive samples with the same label as \(i\)-th index, and $\left| S(\cdot) \right|$ denotes its cardinality.
Specifically, in \emph{one view} settings of GaitGraph2~\cite{teepe2022towards}, set \(I\) contains \(N\) elements, while in \emph{two view} settings of GaitGraph~\cite{teepe2021gaitgraph}, it contains \(2N\) elements. The additional \(N\)  elements are generated by different data augmentations.

\section{Benchmark Results}
\subsection{Overall Experimental Settings}
The initial learning rates for GaitGraph, GaitGraph2, and GaitTR are 0.01, 0.05, and 0.001, respectively, following the OneCycle~\cite{smith2019super} learning rate scheduler.
GaitGraph2 on CASIA-B and Gait3D employ AdamW~\cite{DBLP:conf/iclr/LoshchilovH19} and others utilize Adam~\cite{DBLP:journals/corr/KingmaB14} as the optimizer.
All the models are trained and evaluated on four TITAN Xp 12G GPUs. More details about the sampler, batch size and model structure can be seen in the following sections.
\begin{table}[!h]
    \begin{center}
        \captionsetup{ font=small}
        \caption{The batch size configurations in Vanilla Version.}
        \fontsize{10}{14.5}\selectfont
        \label{BT-vanilla}
        \resizebox{0.47\textwidth}{!}{
            \begin{tabular}{c|cccc}
                \hline
                Model                               & CASIA-B & OUMVLP-Pose & GREW   & Gait3D \\ \hline
                GaitGraph~\cite{teepe2021gaitgraph} & 128     & 128         & 128    & 128    \\
                GaitGraph2~\cite{teepe2022towards}  & 768     & 768         & 768    & 768    \\
                GaitTR~\cite{zhang2022spatial}      & (4,64)  & (32,16)     & (32,8) & (32,4) \\ \hline
            \end{tabular}
        }
    \end{center}
    \vspace{-0.3cm}
\end{table}

\begin{table*}[!h]
    \begin{center}
        \setlength{\abovecaptionskip}{2pt}
        \captionsetup{ font=small}
        \caption{The data augmentation configurations in Vanilla Version.}
        \fontsize{9}{13}\selectfont
        \label{DA-vanilla}
        \resizebox{\textwidth}{!}{
            \begin{tabular}{c|ccccc|cc}
                \hline
                                                    & \multicolumn{5}{c|}{Spatial} & \multicolumn{2}{c}{Sequential}                                                                              \\ \hline
                Model                               & InversePosesPre              & MirrorPoses                    & PointNoise & JointNoise & RandomMove & RandomSelectSequence & FlipSequence \\ \hline
                GaitGraph~\cite{teepe2021gaitgraph} &                              & \checkmark                     & \checkmark & \checkmark &            & \checkmark           & \checkmark   \\
                GaitGraph2~\cite{teepe2022towards}  & \checkmark                   &                                & \checkmark & \checkmark & \checkmark & \checkmark           & \checkmark   \\
                GaitTR ~\cite{zhang2022spatial}     & \checkmark                   &                                & \checkmark & \checkmark &            & \checkmark           &              \\ \hline
            \end{tabular}
        }
    \end{center}
\end{table*}
% CASIA-B
\begin{table*}[!h]
    \begin{center}
        \setlength{\abovecaptionskip}{2pt}
        \captionsetup{ font=small}
        \caption{Rank-1 (\%) performance of three methods on CASIA-B. The results in parentheses are mentioned in the papers.}
        \fontsize{9}{12}\selectfont
        \label{casiab}

        \resizebox{\linewidth}{!}{
            \begin{tabular}{cc|cccc|cccc}
                \hline
                \multicolumn{2}{c|}{Version}                             & \multicolumn{4}{c|}{Vanilla} & \multicolumn{4}{c}{Improved }                                                                                 \\ \hline
                \multicolumn{1}{c|}{Model}                               & Pose Estimator               & NM                            & BG            & CL            & Mean          & NM    & BG    & CL    & Mean  \\ \hline
                \multicolumn{1}{c|}{GaitGraph~\cite{teepe2021gaitgraph}} & HRNet~\cite{8953615}         & 86.37 (87.70)                 & 76.50 (74.80) & 65.24 (66.30) & 76.04 (76.27) & 88.47 & 77.52 & 67.95 & 77.98 \\
                \multicolumn{1}{c|}{GaitGraph2~\cite{teepe2022towards}}  & HRNet~\cite{8953615}         & 80.29 (82.00)                 & 71.40 (73.20) & 63.80 (63.60) & 71.83 (72.93) & 83.60 & 72.80 & 67.01 & 74.47 \\
                \multicolumn{1}{c|}{GaitTR~\cite{zhang2022spatial}}      & HRNet~\cite{8953615}         & 94.72 & 89.29 & 86.65 & 90.22 & 95.55 & 89.79 & 85.76 & 90.37 \\
                \multicolumn{1}{c|}{GaitTR~\cite{zhang2022spatial}}      & SimCC~\cite{li2022simcc}     & 94.91 (96.00)                 & 88.82 (91.30) & 90.34 (90.00) & 91.35 (92.40) & 95.02 & 90.70 & 89.67 & 91.80 \\ \hline
            \end{tabular}
        }
    \end{center}
\end{table*}

% OUMVLP-Pose
\begin{table*}[!h]
    \begin{center}
        \setlength{\abovecaptionskip}{2pt}
        \captionsetup{ font=small}
        \caption{Rank-1 (\%) performance of three methods on OUMVLP-Pose estimated by AlphaPose~\cite{alphapose}, excluding the identical-view cases.}
        \fontsize{9}{12}\selectfont
        \label{oumvlp-pose}

        \begin{tabular}{r|cccccc}
            \hline
            \multirow{3}{*}{Probe} & \multicolumn{6}{c}{Gallery(0\degree  - 270\degree )}                                                                                                                                                                                                        \\ \cline{2-7}
                                   & \multicolumn{3}{c|}{Vanilla}                         & \multicolumn{3}{c}{Improved }                                                                                                                                                                        \\
                                   & GaitGraph~\cite{teepe2021gaitgraph}                  & GaitGraph2~\cite{teepe2022towards} & \multicolumn{1}{c|}{GaitTR~\cite{zhang2022spatial}} & GaitGraph~\cite{teepe2021gaitgraph} & GaitGraph2~\cite{teepe2022towards} & GaitTR~\cite{zhang2022spatial} \\ \hline
            0\degree               & 1.88                                                 & 53.19                              & \multicolumn{1}{c|}{29.59}                          & 41.27                               & 54.88                              & 33.04                          \\
            15\degree              & 3.26                                                 & 66.56                              & \multicolumn{1}{c|}{43.22}                          & 53.19                               & 67.55                              & 46.64                          \\
            30\degree              & 4.01                                                 & 71.04                              & \multicolumn{1}{c|}{47.77}                          & 57.69                               & 71.98                              & 50.96                          \\
            45\degree              & 4.56                                                 & 72.03                              & \multicolumn{1}{c|}{51.45}                          & 59.26                               & 73.80                              & 53.61                          \\
            60\degree              & 3.66                                                 & 67.10                              & \multicolumn{1}{c|}{50.81}                          & 58.19                               & 70.80                              & 53.41                          \\
            75\degree              & 3.25                                                 & 64.90                              & \multicolumn{1}{c|}{45.80}                          & 58.49                               & 68.74                              & 48.99                          \\
            90\degree              & 2.31                                                 & 62.35                              & \multicolumn{1}{c|}{39.47}                          & 55.18                               & 65.62                              & 42.99                          \\
            180\degree             & 1.88                                                 & 47.71                              & \multicolumn{1}{c|}{25.56}                          & 38.62                               & 49.51                              & 29.38                          \\
            195\degree             & 2.69                                                 & 56.41                              & \multicolumn{1}{c|}{33.50}                          & 43.99                               & 58.71                              & 37.29                          \\
            210\degree             & 1.83                                                 & 52.74                              & \multicolumn{1}{c|}{30.48}                          & 42.09                               & 54.29                              & 34.78                          \\
            225\degree             & 3.08                                                 & 67.41                              & \multicolumn{1}{c|}{46.10}                          & 54.45                               & 69.70                              & 48.49                          \\
            240\degree             & 2.60                                                 & 65.27                              & \multicolumn{1}{c|}{45.01}                          & 53.42                               & 67.55                              & 48.19                          \\
            255\degree             & 2.30                                                 & 63.93                              & \multicolumn{1}{c|}{40.05}                          & 52.47                               & 66.98                              & 44.64                          \\
            270\degree             & 2.02                                                 & 58.91                              & \multicolumn{1}{c|}{35.47}                          & 49.00                               & 63.28                              & 38.09                          \\ \hline
            Mean                   & 2.81                                                 & 62.11                              & \multicolumn{1}{c|}{40.30}                          & 51.24                               & 64.53                              & 43.61                          \\ \hline
        \end{tabular}
    \end{center}
\end{table*}
% GREW
\begin{table*}[!h]
    \begin{center}
        \setlength{\abovecaptionskip}{2pt}
        \captionsetup{ font=small}
        \caption{Rank-1 (\%), Rank-5 (\%), Rank-10 (\%) performance of three methods on GREW.}
        \label{grew}
        \fontsize{9}{12}\selectfont
        \begin{tabular}{cc|ccc|ccc}
            \hline
            \multicolumn{2}{c|}{Version}                             & \multicolumn{3}{c|}{Vanilla} & \multicolumn{3}{c}{Improved }                                                \\ \hline
            \multicolumn{1}{c|}{Model}                               & Pose Estimator               & Rank-1                        & Rank-5 & Rank-10 & Rank-1 & Rank-5 & Rank-10 \\ \hline
            \multicolumn{1}{c|}{GaitGraph~\cite{teepe2021gaitgraph}} & HRNet~\cite{8953615}         & 10.18                         & 19.57  & 24.73   & 36.08  & 56.69  & 64.89   \\
            \multicolumn{1}{c|}{GaitGraph2~\cite{teepe2022towards}}  & HRNet~\cite{8953615}         & 34.78                         & 49.69  & 55.51   & 44.41  & 59.04  & 64.69   \\
            \multicolumn{1}{c|}{GaitTR~\cite{zhang2022spatial}}      & HRNet~\cite{8953615}         & 48.58                         & 65.48  & 71.58   & 55.33  & 71.40  & 76.78   \\ \hline
        \end{tabular}
    \end{center}
\end{table*}

\begin{table*}[!h]
    \begin{center}
        \setlength{\abovecaptionskip}{2pt}
        \captionsetup{ font=small}
        \caption{Rank-1 (\%), Rank-5 (\%), Rank-10 (\%) performance of three methods on Gait3D.}
        \label{gait3d}
        \fontsize{9}{12}\selectfont
        \begin{tabular}{cc|ccc|ccc}
            \hline
            \multicolumn{2}{c|}{Version}                             & \multicolumn{3}{c|}{Vanilla} & \multicolumn{3}{c}{Improved }                                                \\ \hline
            \multicolumn{1}{c|}{Model}                               & Pose Estimator               & Rank-1                        & Rank-5 & Rank-10 & Rank-1 & Rank-5 & Rank-10 \\ \hline
            \multicolumn{1}{c|}{GaitGraph~\cite{teepe2021gaitgraph}} & HRNet~\cite{8953615}         & 8.6                           & 17.3   & 23.8    & 14.6   & 31.3   & 38.8    \\
            \multicolumn{1}{c|}{GaitGraph2~\cite{teepe2022towards}}  & HRNet~\cite{8953615}         & 11.2                          & 24.0   & 31.2    & 12.5   & 24.7   & 30.6    \\
            \multicolumn{1}{c|}{GaitTR~\cite{zhang2022spatial}}      & HRNet~\cite{8953615}         & 7.2                           & 15.5   & 20.5    & 9.7    & 21.8   & 28.4    \\ \hline
        \end{tabular}
    \end{center}
    \vspace{-0.3cm}
\end{table*}
%------------------

\subsection{Vanilla Version}

For Vanilla Version, our goal is to maintain consistency with the original codebases or descriptions in the papers to the greatest extent possible. This involves preserving identical types of model structures, data augmentations, samplers, and batch sizes described in the original papers~\cite{teepe2021gaitgraph,teepe2022towards,zhang2022spatial}.
\textbf{(1)} Data Augmentations. As shown in \cref{DA-vanilla}, we employ the same types of data augmentations with original codebases and papers in Vanilla Version.
\textbf{(2)} Samplers. Random Sampler is adopted in GaitGraph and GaitGraph2, while GaitTR utilizes Triplet Sampler.
\textbf{(3)} Batch Sizes. Three models utilize different batch sizes as described in the papers, shown in \cref{BT-vanilla}.

The results can be seen in~\cref{casiab,oumvlp-pose,grew,gait3d}. We achieve satisfactory results on CASIA-B in comparison to the results mentioned in the papers, with a maximum deviation of less than 2\% in terms of Rank-1 accuracy. On OUMVLP-Pose, GREW and Gait3D, the reproduced algorithms still yield a relatively stable performance across indoor and outdoor scenarios and varying dataset scales, especially for GaitGraph2 and GaitTR.

\subsection{Improved Version}
\begin{figure}
    \centering
    \includegraphics[width=\linewidth]{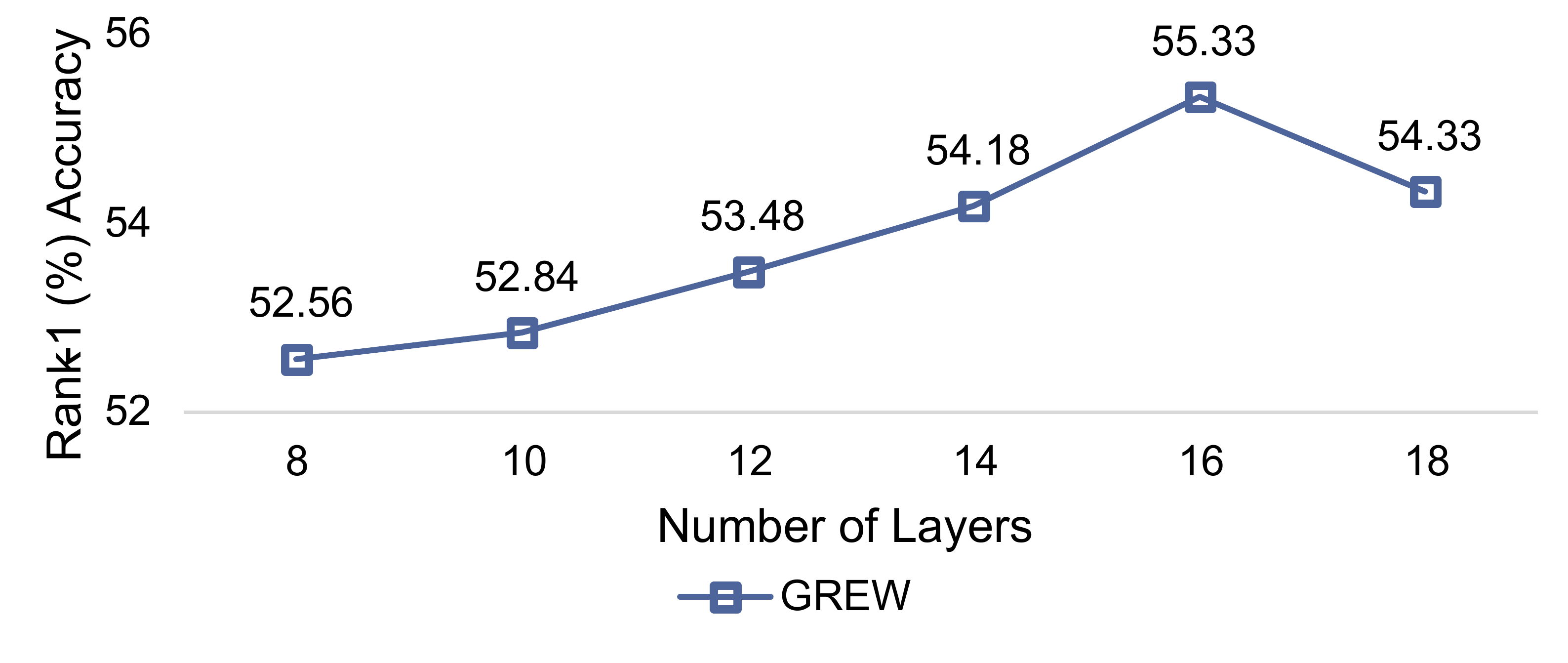}
    \includegraphics[width=\linewidth]{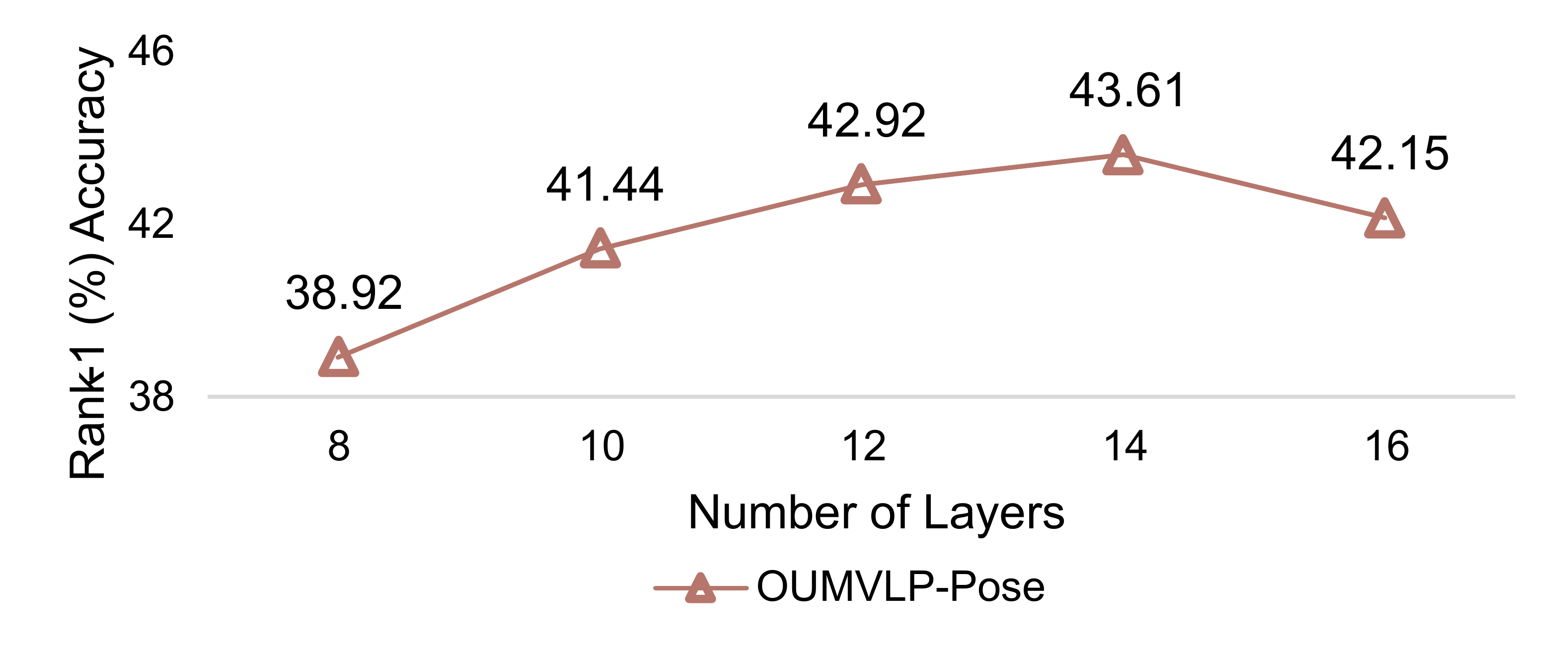}
    \caption{Impact of layer numbers on GaitTR.}
    \label{fig:line_chart}
    \vspace{-0.3cm}
\end{figure}
Although Vanilla Version of the three methods demonstrates satisfactory performance, there is still plenty of room for improvement:
\textbf{(1)} As shown in~\cref{casiab,oumvlp-pose,grew,gait3d}, some results of GaitGraph are drastically inconsistent on untested benchmarks, \eg, Rank-1 accuracy of only 2.81\% on OUMVLP-Pose and 10.18\% on GREW.
\textbf{(2)} Pose-based gait recognition is sensitive to data augmentations, and thus original settings can be unsuitable for datasets with diverse scenarios.
\textbf{(3)} Deep network structures have been proved to be effective in both deep learning~\cite{he2016deep} and appearance-based gait recognition~\cite{fan2023exploring}. Therefore, a deeper network is necessary to achieve optimal performance for larger-scale datasets.
Taking these factors into account, we have introduced an Improved Version of FastPoseGait to fully unleash the model's potential on all four benchmarks. This version incorporates a more unified experimental setup and provides a comprehensive analysis of samplers, data augmentations, and model structures.

\begin{table}[]
    \begin{center}
        \captionsetup{ font=small}
        \caption{The impact of different sampler strategies in GaitGraph on the OUMVLP-Pose dataset.}
        \label{sampler}
        \vspace{-0.2cm}
        \fontsize{7}{9.5}\selectfont
        \resizebox{0.38\textwidth}{!}{
            \begin{tabular}{c|c|c}
                \hline
                Sampler         & Batch Size       & Rank-1         \\ \hline
                Random Sampler  & 128              & 2.81           \\
                Triplet Sampler & (32,16)          & 49.68          \\
                Triplet Sampler & \textbf{(256,2)} & \textbf{51.24} \\ \hline
            \end{tabular}
        }
    \end{center}
    \vspace{-0.6cm}
\end{table}

\textbf{(1) Sampler.}
To tackle the issue of selecting positive pairs in GaitGraph and GaitGraph2, we have replaced Random Sampler with Triplet Sampler. This modification prevents the supervised contrastive loss from degrading to an unsupervised one, maintaining the integrity of the loss function's gradient structure. Additionally, to address overfitting on CASIA-B for GaitGraph, we utilize the Random Triplet Sampler described in Section~\ref{section:3.1}. As shown in the preliminary study of \cref{sampler}, the performance of GaitGraph on OUMVLP-Pose can be improved  by up to 48.4\%.

\textbf{(2) Data Augmentations.}
For pose-based gait recognition, appropriate data augmentation holds significant importance. We have extensively experimented with various data augmentations and explored different combinations on both indoor and outdoor benchmarks. The details of these experiments can be seen in \cref{DA-casiab-GaitGraph,DA-casiab-gaitgraph2,DA-casiab-gaittr,DA-gait3d-gaitgraph,DA-gait3d-gaitgraph2,DA-gait3d-gaittr}. Furthermore, due to the observed tendency of methods to overfit on CASIA-B, we have applied the same augmentations used on Gait3D to GREW and OUMVLP-Pose datasets. The maximum improvement achieved through data augmentations is 15.1\%.

\textbf{(3) Network Depth.}
In the case of GaitGraph and GaitGraph2, we adhere to the original network depth
settings. Since GaitGraph2, as an enhanced version of GaitGraph, adjusts the model layers to accommodate different dataset scales. To determine the optimal number of layers of GaitTR, we examine the network depth specifically on the GREW and OUMVLP-Pose. As shown in \cref{fig:line_chart}, GaitTR achieves the highest performance of 16 layers on GREW and 14 layers on OUMVLP-Pose with the improvement of 2.8\% and 4.7\%, respectively.

Compared with Vanilla Version, the overall performance of the Improved Version witnesses significant enhancements shown in~\cref{casiab,oumvlp-pose,grew,gait3d}, \ie, +1.9\%, +48.4\%, +25.9\% and +6.0\% for GaitGraph, +2.6\%, +2.4\%, +9.6\% and +1.3\% for GaitGraph2, +0.5\%, +3.3\%, +6.8\% and +2.5\% for GaitTR on CASIA-B, OUMVLP-Pose, GREW and Gait3D, respectively. The Improved Version offers comparable experimental results for pose-based methods across various datasets, showcasing both unleashed potentials of the models and effectiveness of pose-based approaches.

%-------------------------------------------------------------------------
\section{Conclusion}

FastPoseGait is an open-source toolbox designed for pose-based gait recognition. Unlike other projects that focus on a single algorithm, FastPoseGait integrates three representative algorithms into a unified framework with a highly modular structure, which allows for easy comparison of effectiveness and efficiency. The toolbox encompasses diverse benchmarks, pre-trained models, and comprehensive results, providing valuable resources for further advancements in pose-based gait recognition research.

\begin{table*}[!h]
    \begin{center}
        \captionsetup{ font=small}
        \caption{The data augmentation study of  GaitGraph on CASIA-B.}
        \fontsize{10}{13}\selectfont
        \vspace{-0.2cm}
        \label{DA-casiab-GaitGraph}
        \begin{tabular}{ccccc|c|cccc}
            \hline
            \multicolumn{5}{c|}{Spatial} & Sequential  & \multicolumn{4}{c}{CASIA-B}                                                                                                              \\ \hline
            InversePosesPre              & MirrorPoses & PointNoise                  & JointNoise & RandomMove & FlipSequence & NM             & BG             & CL             & Mean           \\ \hline
            —                            & —           & —                           & —          & —          & —            & 73.75          & 63.19          & 51.62          & 62.85          \\
                                         &             &                             &            &            & \checkmark   & \textbf{74.32} & \textbf{63.95} & \textbf{52.05} & \textbf{63.44} \\
            \checkmark                   &             &                             &            &            & \checkmark   & 73.16          & 59.96          & 43.53          & 58.88          \\
                                         & \checkmark  &                             &            &            & \checkmark   & 71.63          & 58.87          & 42.37          & 57.62          \\
                                         &             & \checkmark                  &            &            & \checkmark   & \textbf{85.74} & \textbf{71.54} & \textbf{61.78} & \textbf{73.02} \\
                                         &             &                             & \checkmark &            & \checkmark   & \textbf{75.77} & \textbf{64.88} & \textbf{49.66} & \textbf{63.44} \\
                                         &             &                             &            & \checkmark & \checkmark   & 68.56          & 54.96          & 39.04          & 54.19          \\ \hline
            \checkmark                   &             & \checkmark                  &            &            & \checkmark   & 84.85          & 73.53          & 62.28          & 73.55          \\
            \checkmark                   &             &                             & \checkmark &            & \checkmark   & 77.09          & 67.79          & 53.08          & 65.99          \\
                                         &             & \checkmark                  & \checkmark &            & \checkmark   & 85.98          & 74.22          & 64.85          & 75.02          \\\rowcolor{gray!30}
            \checkmark                   &             & \checkmark                  & \checkmark &            & \checkmark   & \textbf{88.47} & \textbf{77.52} & \textbf{67.95} & \textbf{77.98} \\ \hline
        \end{tabular}
    \end{center}
\end{table*}

\begin{table*}[!h]
    \begin{center}
        \captionsetup{ font=small}
        \caption{The data augmentation study of  GaitGraph2 on CASIA-B.}
        \label{DA-casiab-gaitgraph2}
        \fontsize{10}{13}\selectfont
        \vspace{-0.2cm}
        \begin{tabular}{ccccc|c|cccc}
            \hline
            \multicolumn{5}{c|}{Spatial} & Sequential  & \multicolumn{4}{c}{CASIA-B}                                                                                                                        \\ \hline
            InversePosesPre              & MirrorPoses & PointNoise                  & JointNoise     & RandomMove     & FlipSequence   & NM             & BG             & CL             & Mean           \\ \hline
            —                            & —           & —                           & —              & —              & —              & 78.76          & 67.89          & 60.18          & 68.94          \\
            \checkmark                   &             &                             &                &                &                & \textbf{81.85} & \textbf{69.53} & \textbf{62.06} & \textbf{71.15} \\
                                         & \checkmark  &                             &                &                &                & \textbf{82.33} & \textbf{71.21} & \textbf{63.06} & \textbf{72.20} \\ \rowcolor{gray!30}
                                         &             & \textbf{\checkmark}         &                &                &                & \textbf{83.60} & \textbf{72.80} & \textbf{67.01} & \textbf{74.47} \\
                                         &             &                             & \checkmark     &                &                & \textbf{78.39} & \textbf{68.69} & \textbf{63.22} & \textbf{70.10} \\
                                         &             &                             &                & \checkmark     &                & \textbf{81.08} & \textbf{68.61} & \textbf{61.14} & \textbf{70.28} \\
                                         &             &                             &                &                & \checkmark     & \textbf{82.21} & \textbf{70.57} & \textbf{60.19} & \textbf{70.99} \\ \hline
                                         & \checkmark  & \checkmark                  &                &                &                & 81.81          & 68.82          & 63.64          & 71.42          \\
                                         &
                                         & \checkmark  &
                                         & \checkmark  & \textbf{}                   & \textbf{84.31} & \textbf{71.96} & \textbf{65.78} & \textbf{74.02}                                                    \\
                                         &             & \checkmark                  &                &                & \checkmark     & 81.04          & 69.33          & 60.16          & 70.18          \\
            \checkmark                   &             & \checkmark                  &                &                &                & 81.40          & 69.70          & 62.55          & 71.22          \\
                                         &             & \checkmark                  & \checkmark     &                &                & 81.40          & 69.48          & 64.12          & 71.67          \\
                                         & \checkmark  & \checkmark                  &                &                & \checkmark     & 84.31          & 72.27          & 63.20          & 73.26          \\
                                         &             & \checkmark                  & \checkmark     &                & \checkmark     & 82.62          & 70.87          & 62.88          & 72.12          \\
                                         & \checkmark  & \checkmark                  & \checkmark     & \checkmark     & \checkmark     & 84.48          & 72.75          & 64.54          & 73.92          \\
                                         &             & \checkmark                  & \checkmark     & \checkmark     & \checkmark     & 83.16          & 71.49          & 62.26          & 72.31          \\ \hline
        \end{tabular}
    \end{center}
\end{table*}

% CASIA-B  GaitTR
%-------------
\begin{table*}[!h]
    \begin{center}
        \captionsetup{ font=small}
        \caption{The data augmentation study of  GaitTR on CASIA-B.}
        \label{DA-casiab-gaittr}
        \fontsize{10}{13}\selectfont
        \vspace{-0.2cm}
        \begin{tabular}{ccccc|c|cccc}
            \hline
            \multicolumn{5}{c|}{Spatial} & Sequential          & \multicolumn{4}{c}{CASIA-B}                                                                                                                       \\ \hline
            InversePosesPre              & MirrorPoses         & PointNoise                  & JointNoise          & RandomMove & FlipSequence & NM             & BG             & CL             & Mean           \\ \hline
            —                            & —                   & —                           & —                   & —          & —            & 92.75          & 88.78          & 88.41          & 89.98          \\
            \checkmark                   &                     &                             &                     &            &              & \textbf{94.31} & \textbf{89.68} & \textbf{88.97} & \textbf{90.99} \\
                                         & \checkmark          &                             &                     &            &              & \textbf{94.03} & \textbf{90.04} & \textbf{88.34} & \textbf{90.80} \\
                                         &                     & \checkmark                  &                     &            &              & 93.49          & 88.48          & 87.23          & 89.73          \\
                                         &                     &                             & \checkmark          &            &              & \textbf{93.95} & \textbf{88.88} & \textbf{89.72} & \textbf{90.85} \\
                                         &                     &                             &                     & \checkmark &              & 92.84          & 87.78          & 86.16          & 88.93          \\
                                         &                     &                             &                     &            & \checkmark   & 93.21          & 88.87          & 86.80          & 89.63          \\
            \hline  \rowcolor{gray!30}
            \textbf{\checkmark}          & \textbf{\checkmark} &                             & \textbf{\checkmark} &            &              & \textbf{95.02} & \textbf{90.70} & \textbf{89.67} & \textbf{91.80} \\ \hline
        \end{tabular}
    \end{center}
\end{table*}

%-------------

% Gait3D  GaitGraph
\begin{table*}[!h]
    \captionsetup{ font=small}
    \caption{The data augmentation study of  GaitGraph on Gait3D.}
    \label{DA-gait3d-gaitgraph}
    \fontsize{10}{13}\selectfont
    \vspace{-0.2cm}
    \begin{tabular}{ccccc|c|ccc}
        \hline
        \multicolumn{5}{c|}{Spatial} & \multicolumn{1}{c|}{Sequential} & \multicolumn{3}{c}{Gait3D}                                                                                          \\ \hline
        InversePosesPre              & MirrorPoses                     & PointNoise                 & JointNoise & RandomMove & FlipSequence & Rank-1        & Rank-5        & Rank-10       \\ \hline
        —                            & —                               & —                          & —          & —          & —            & 11.4          & 24.2          & 32.8          \\
        \checkmark                   &                                 &                            &            &            &              & 10.3          & 23.6          & 31.3          \\
                                     & \checkmark                      &                            &            &            &              & \textbf{11.1} & \textbf{25.2} & \textbf{33.0} \\
                                     &                                 & \checkmark                 &            &            &              & \textbf{12.1} & \textbf{28.1} & \textbf{36.4} \\
                                     &                                 &                            & \checkmark &            &              & \textbf{10.7} & \textbf{25.9} & \textbf{33.0} \\
                                     &                                 &                            &            & \checkmark &              & 10.3          & 24.8          & 32.2          \\
                                     &                                 &                            &            &            & \checkmark   & \textbf{11.0} & \textbf{26.0} & \textbf{33.7} \\
        \hline
        \textbf{}                    & \checkmark                      & \checkmark                 & \textbf{}  & \textbf{}  & \textbf{}    & 13.2          & 29.0          & 38.0          \\
        \rowcolor{gray!30}
                                     & \checkmark                      & \checkmark                 & \checkmark &            &              & \textbf{14.6} & \textbf{31.3} & \textbf{38.8} \\
                                     & \checkmark                      & \checkmark                 & \checkmark &            & \checkmark   & 13.2          & 28.2          & 35.5          \\
        \hline
    \end{tabular}
\end{table*}
% Gait3D  GaitGraph2
%-------
\begin{table*}[!h]
    \begin{center}
        \captionsetup{ font=small}
        \caption{The data augmentation study of  GaitGraph2 on Gait3D.}
        \label{DA-gait3d-gaitgraph2}
        \fontsize{10}{13}\selectfont
        \vspace{-0.2cm}
        \begin{tabular}{ccccc|c|ccc}
            \hline
            \multicolumn{5}{c|}{Spatial} & \multicolumn{1}{c|}{Sequential} & \multicolumn{3}{c}{Gait3D}                                                                                          \\ \hline
            InversePosesPre              & MirrorPoses                     & PointNoise                 & JointNoise & RandomMove & FlipSequence & Rank-1        & Rank-5        & Rank-10       \\ \hline
            —                            & —                               & —                          & —          & —          & —            & 11.3          & 24.1          & 31.1          \\
            \checkmark                   &                                 &                            &            &            &              & \textbf{11.4} & \textbf{24.4} & \textbf{33.9} \\
                                         & \checkmark                      &                            &            &            &              & 10.5          & 24.4          & 32.6          \\
                                         &                                 & \checkmark                 &            &            &              & \textbf{11.9} & \textbf{26.1} & \textbf{32.8} \\
                                         &                                 &                            & \checkmark &            &              & 10.3          & 23.1          & 31.8          \\
                                         &                                 &                            &            & \checkmark &              & 10.8          & 24.5          & 32.0          \\
                                         &                                 &                            &            &            & \checkmark   & \textbf{11.7} & \textbf{25.9} & \textbf{33.0} \\ \hline
            \checkmark                   &                                 & \checkmark                 &            &            & \checkmark   & 12.6          & 23.4          & 29.9          \\\rowcolor{gray!30}
                                         &                                 & \checkmark                 &            &            & \checkmark   & \textbf{12.5} & \textbf{24.7} & \textbf{30.6} \\ \hline
        \end{tabular}
    \end{center}
\end{table*}
%-------

% Gait3D  GaitTR

\begin{table*}[!h]
    \begin{center}
        \captionsetup{ font=small}
        \caption{The data augmentation study of  GaitTR on Gait3D.}
        \label{DA-gait3d-gaittr}
        \fontsize{10}{13}\selectfont
        \vspace{-0.2cm}
        \begin{tabular}{ccccc|c|ccc}
            \hline
            \multicolumn{5}{c|}{Spatial} & Sequential  & \multicolumn{3}{c}{Gait3D}                                                                                         \\ \hline
            InversePosesPre              & MirrorPoses & PointNoise                 & JointNoise & RandomMove & FlipSequence & Rank-1       & Rank-5        & Rank-10       \\ \hline
            —                            & —           & —                          & —          & —          & —            & 4.0          & 12.9          & 18.0          \\
            \checkmark                   &             &                            &            &            &              & \textbf{7.8} & \textbf{15.4} & \textbf{21.2} \\
                                         & \checkmark  &                            &            &            &              & 4.3          & 12.6          & 18.2          \\
                                         &             & \checkmark                 &            &            &              & 4.7          & 12.5          & 16.8          \\
                                         &             &                            & \checkmark &            &              & 4.4          & 13.2          & 17.6          \\
                                         &             &                            &            & \checkmark &              & 3.7          & 11.9          & 17.2          \\\rowcolor{gray!30}
                                         &             &                            &            &            & \checkmark   & \textbf{9.7} & \textbf{21.8} & \textbf{28.4} \\
            \hline
            \checkmark                   &             &                            &            &            & \checkmark   & \textbf{8.6} & \textbf{19.2} & \textbf{26.0} \\
            \hline
        \end{tabular}
    \end{center}
\end{table*}

\clearpage
\clearpage

{\small
    \bibliographystyle{ieee_fullname}
    \bibliography{egbib}
}

\end{document}